\documentclass[11pt]{article}

\usepackage[margin=1in]{geometry}
\usepackage{microtype}
\usepackage[round,authoryear]{natbib}
\usepackage{hyperref}
\usepackage{url}
\usepackage{booktabs}
\usepackage{amsmath}
\usepackage{amssymb}
\usepackage{multirow}
\usepackage{float}
\usepackage{graphicx}
\usepackage{xcolor}

\definecolor{darkblue}{rgb}{0, 0, 0.5}
\hypersetup{colorlinks=true, citecolor=darkblue, linkcolor=darkblue, urlcolor=darkblue}
\title{Sensitivity-Aware Thresholding and Token Routing\\ for Activation Sparsification in Large Language Models}
\author{
Bishmoy Paul\\
Santa Clara University \\
\texttt{bishmoypaul.contact@gmail.com}
\and
Youngmin Yi\\
Sogang University \\
\texttt{ymyi@sogang.ac.kr}
\and
Hoeseok Yang\\
Santa Clara University\\
\texttt{hoeseok.yang@scu.edu}
}
\date{}

\begin{document}

\maketitle

\begin{abstract}
Efficient inference in Large Language Models (LLMs) requires deciding where computation can be reduced while preserving model quality. We study this problem through multilayer perceptron (MLP) activation sparsification and token-level conditional routing. We first propose Sensitivity-Aware Thresholding for Sparsity (SATS), a threshold calibration method to choose layerwise gate thresholds using a local MLP output sensitivity proxy rather than calibrating thresholds directly from activation percentiles. While SATS retains the existing mechanism of sparsifying MLP activations by thresholding gate activations, it replaces percentile-based calibration with a sensitivity-aware selection rule. We then introduce a lightweight token routing framework that dynamically selects between a base path and a modified path on a per-token basis, rather than applying the modified computation uniformly to all tokens. We evaluate both methods on 
multiple recent open-weight LLMs.
Our results show that SATS improves over the threshold-based sparsification baseline at matched actual sparsity
and that token routing yields a more favorable quality-throughput trade-off than static activation modification baselines. 
Overall, our results suggest that improved threshold calibration and token routing can improve the quality-throughput trade-off in LLMs.
\end{abstract}

\section{Introduction}
\label{sec:Intro}
Large Language Models (LLMs) are widely used across various tasks, but their inference cost remains a major bottleneck. A significant portion of this cost comes from the multilayer perceptron (MLP) blocks inside each transformer layer. Therefore, optimizing MLP blocks can speed up on-device LLM inference. Recent works~\citep{cats,relu_squared} have shown that these activations contain significant sparsity that can be exploited by thresholding MLP gate activations, which can speed up LLM inference without modifying the overall transformer architecture. However, deciding which activations to remove and when to apply the modified computation remains a challenge.

A common strategy to address this challenge is to calibrate one activation threshold per layer from activation statistics from a calibration dataset. CATS \citep{cats} uses a percentile-based approach to choose these layerwise thresholds. But percentile-based calibration focuses primarily on how many activations are removed, instead of examining how much damage a given threshold causes to the resultant MLP output. Two adjacent thresholds in the activation layer can have drastically different amounts of information lost in the final output. Even more broadly, activation modification is applied uniformly across all tokens for a generation prompt, even though the impact of the modification is not necessarily uniform across tokens.

In this work, we study sparsification of the MLP layer in LLMs at two levels. First, we propose \textbf{Sensitivity-aware Thresholding for Sparsity (SATS)}, a threshold calibration method that replaces percentile-based threshold selection with a sensitivity-aware method based on layerwise MLP output distortion. SATS chooses thresholds to satisfy a local layer sensitivity budget, and the layerwise final operating points are chosen to satisfy a global target sparsity. Secondly, we introduce a lightweight \textbf{token routing} method that decides whether to use the dense MLP path or the sparse path (that uses the activation thresholding) for each token. The routing method only requires the token identity, which allows it to be simplified into a look-up table with negligible effect on the inference speed.

We evaluate both of the methods on llama 3.1 8B~\citep{llama3} and Qwen 3 8B~\citep{qwen3}. Across both model families, SATS provides improved quality compared to percentile-based thresholding at matched realized sparsity. We then show that token routing further improves the quality-throughput trade-off relative to static sparse or dense model execution. Our main contributions are as follows:
\begin{itemize}
    \item We propose SATS, a sensitivity-aware threshold calibration method for activation sparsification that preserves the existing runtime activation threshold masking method 
    \item We show that SATS improves over percentile-based threshold calibration at matched realized sparsity on two recent open-weight LLM families. 
    \item We introduce a token-identity-based routing method that selects between the dense and sparse computation paths at inference with minimal routing overhead
    \item We show that token routing improves the quality-throughput trade-off relative to the static sparse execution while remaining faster than dense inference.
\end{itemize}

\section{Background}\label{others}
We focus on the transformer-based LLMs that use Gated MLP blocks, such as llama~\citep{llama} and Qwen~\citep{qwen} families. Given an input vector $x$, the gated MLP block can be represented using the following equation:
\begin{equation}
    \mathrm{Gated\text{-}MLP}(x) := (\mathrm{Activation}(xW_{\mathrm{gate}})*(xW_{\mathrm{up}}))W_{\mathrm{down}}
\end{equation}

where $W_{gate}$, $W_{up}$ and $W_{down}$ represent the weights in the MLP layer. While the activation in the MLP layer can vary, SiLU \citep{silu} is a widely adopted activation, which can be represented using the following equation:

\begin{equation}
    \mathrm{SiLU} := x*\mathrm{sigmoid}(x) = \frac{x}{1+e^{-x}} 
\end{equation}

CATS~\citep{cats} seeks to utilize activation sparsity in the MLP layers by applying specific thresholds to sparsify the values in the activation. Given a sparsity level $k$ (e.g. 50\%) as input, the activations over a random subset of the training data, limited to 500 data points, are calculated. This is used to compute the cutoff threshold as the kth percentile of the resulting values.
\begin{equation}
    t := \min\{t^{\prime} : F(t^{\prime}) \ge k\}
\end{equation}

After calculating the threshold, CATS applies the following activation function on top of the existing activation function in the MLP layer.
\begin{equation}
    \mathrm{CATS}_{t}(x_{j}) := \begin{cases} x_{j}, & \mathrm {if}\ |x_{j}| \ge t \\ 0, & \mathrm{if }\ |x_{j}| < t\end{cases}
\end{equation}

For the input $SiLU(xW_{gate})$, this becomes:
\begin{equation}
    \mathrm{CATS}_{t}(\mathrm{SiLU}(xW_{gate})) := \begin{cases} \mathrm{SiLU}(xW_{gate}), & \mathrm {if }\ |\mathrm{SiLU}(xW_{gate})| \ge t \\ 0, & \mathrm{if }\ |\mathrm{SiLU}(xW_{gate})| < t\end{cases}
\end{equation}

We extend the CATS framework for more damage-aware thresholding. CATS controls the global sparsity level, but does not explicitly account for the damage caused by removing specific neurons. Our proposed Sensitivity-Aware Thresholding for Sparsity (SATS) framework replaces percentile-based sparsity calibration with a sensitivity-aware threshold selection, while preserving the hardware-efficient CATS runtime masking form.

\section{Method}
\label{sec:Method}
\subsection{SATS: Sensitivity-Aware Thresholding for Sparsity}
\label{subsec:SATS}

To quantify the impact of zeroing out neurons under a specific threshold, we define a local MLP distortion metric inspired by the Cumulative Errors of Tail Truncation (CETT) introduced in  $\mathrm{ReLU}^2$~\citep{relu_squared}. Their work demonstrated that a larger MLP truncation error is associated with worse model performance, which motivates our method of using local distortion as a calibration signal. Unlike $\mathrm{ReLU}^2$, which studies sparsification by thresholding MLP outputs directly, we use this signal only to calibrate the per-layer gate thresholds for CATS-style activation masking. 

\begin{figure}
\includegraphics[width=1\textwidth]{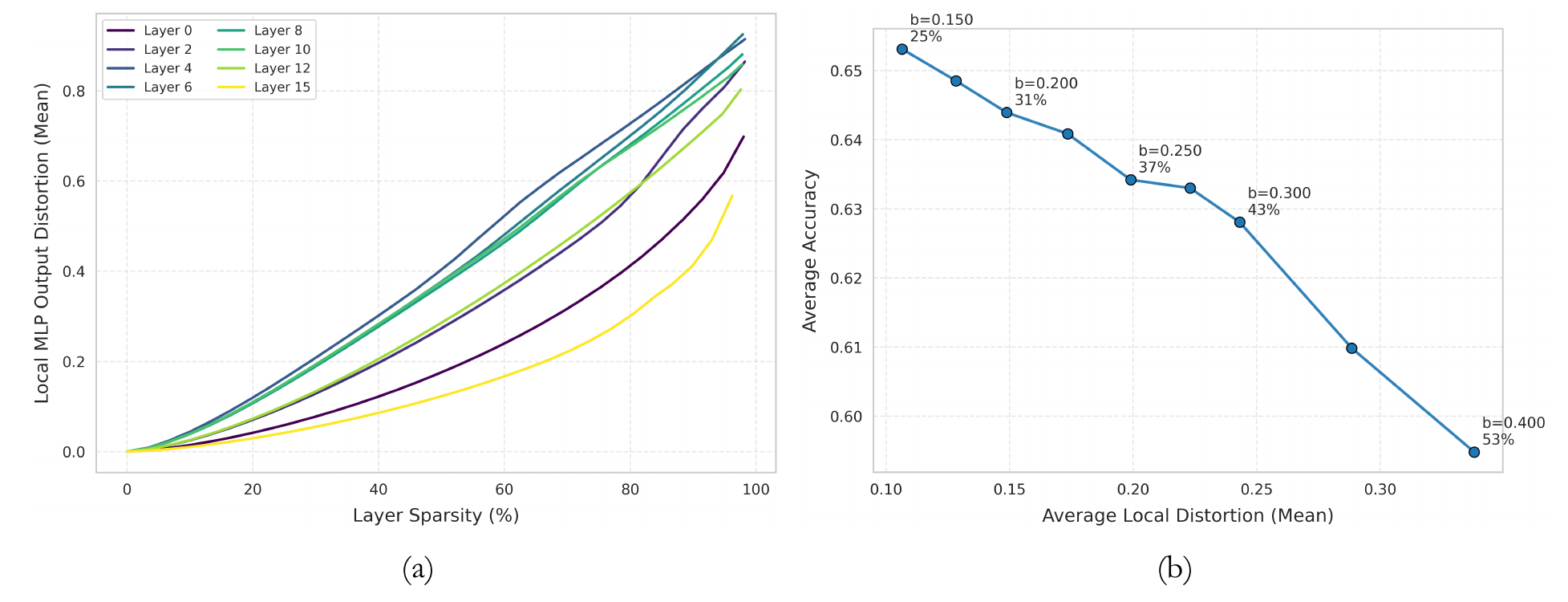}
  \vspace{-5pt}
  \caption{(a) Layerwise distortion under different gate thresholding for llama 3.2 1B. For representative MLP layers, we sweep candidate gate thresholds and measure the layer sparsity and local MLP output distortion on the calibration split. Similar sparsity levels can induce different local distortions across layers (b) Average local distortion vs downstream average accuracy for a SATS budget sweep on llama 3.2 1B. Each point corresponds to a shared layerwise budget $b$ (with its realized sparsity level annotated with the points)}
  \label{fig:sats_budget_sweep}

\end{figure}

The CATS method selects thresholds from the empirical distribution of post-gate activations, which directly models sparsity but does not model the distortion caused by thresholding. Applying adjacent thresholds on the gate layer can contribute differently to the final MLP output depending on the up-projection value and the down-projection weights. As shown in Figure \ref{fig:sats_budget_sweep} (a), similar layer sparsity can introduce substantially different local distortion across layers. Since the sensitivity of the layers to distortion (distortion measured using Equation \ref{eq:_err}) due to thresholding can vary widely, we replace the percentile-based threshold calibration with a sensitivity-aware calibration method while preserving the CATS runtime masking method.

For a candidate threshold $t$, let $y_\ell(x)$ denote the dense MLP output at layer $\ell$, and let $\tilde{y}_{\ell,t}(x)$ denote the thresholded MLP output obtained after removing values whose post-gate activation magnitude falls below $t$. We define the local thresholding error as:
\begin{equation}
    \label{eq:_err}
    \mathrm{Err}_\ell(t) = \mathbb{E}_{x}\left[\frac{||y_\ell(x) - \tilde{y}_{\ell,t}(x)||_2}{||y_\ell(x)||_2 + \epsilon}\right]
\end{equation}

where $\epsilon$ is a small constant for numerical stability. This score measures how much thresholding at a specific $t$ distorts the MLP output at layer $\ell$.

As in CATS, we first build a candidate threshold set $\mathcal{T}$ from a calibration dataset using the post-gate activation magnitudes. This maintains the inference-time gate masking method from CATS, however the threshold selection method is modified. For each candidate threshold, we estimate its local error score $\mathrm{Err}_\ell(t)$ on a calibration dataset. 
We then introduce a layer-wise error budget $b$ and for each layer $\ell$, we select the value of the largest threshold $t_\ell$ whose local error does not exceed that budget.

\begin{equation}
    t_\ell^\star(b) = \max\{t \in \mathcal{T}_\ell : \mathrm{Err}_\ell(t) \le b\}
\end{equation}

For any fixed budget $b$, applying this rule independently across all layers yields a model-wide threshold assignment.
\begin{equation}
    T^\star(b) = [t_1^\star(b),\dots,t_L^\star(b)]
\end{equation}

This defines one sparse model for each choice of $b$. Larger budgets permit more aggressive thresholding and generally lead to higher sparsity. Therefore, the choice of $b$ is particularly important here. While $b$ is selected to model the local MLP distortion, it remains a useful signal in modeling the end-to-end LLM performance. Figure \ref{fig:sats_budget_sweep} (b) demonstrates that local MLP distortion tracks end-to-end model performance consistently, which shows that it is informative enough to guide threshold calibration.

We treat $b$ as a calibration parameter used to reach a desired sparsity level. Specifically, for a target sparsity level $s_{\mathrm{tar}}$, each candidate budget $b$ generates a threshold assignment $T^\star(b)$, which yields a realized average sparsity S(b), defined as:
\begin{equation}
    S(b) = \frac{1}{L}\sum_{\ell=1}^{L} \mathrm{sp}_\ell(t_\ell^\star(b))
\end{equation}
where $\mathrm{sp}_\ell(t)$ denotes the expected sparsity at layer $\ell$ under a given threshold $t$. We then select the budget whose realized sparsity is closest to the target:
\begin{equation}
    b^\star = \arg\min_b | S(b) - s_{\mathrm{tar}}|
\end{equation}
This makes the roles of $b$ and $s_{\mathrm{tar}}$ distinct. Budget $b$ controls which threshold is admissible under the local error criterion, and target sparsity $s_{\mathrm{tar}}$ is used only in an outer calibration loop to choose among the resulting sparse models. 

In practice, we implement this calibration with a coarse-to-fine search. We first sweep a coarse set of budget values and identify the region whose realized sparsity is closest to our target sparsity. We then refine locally around that region by searching nearby threshold candidates more densely for each layer. Conceptually, this approach recasts threshold calibration as a target-sparsity search over a sensitivity-constrained threshold selection. 

\subsection{Token Routing}
\label{subsec:Token_Route}

\begin{figure}
    \centering
\includegraphics[width=1\linewidth]{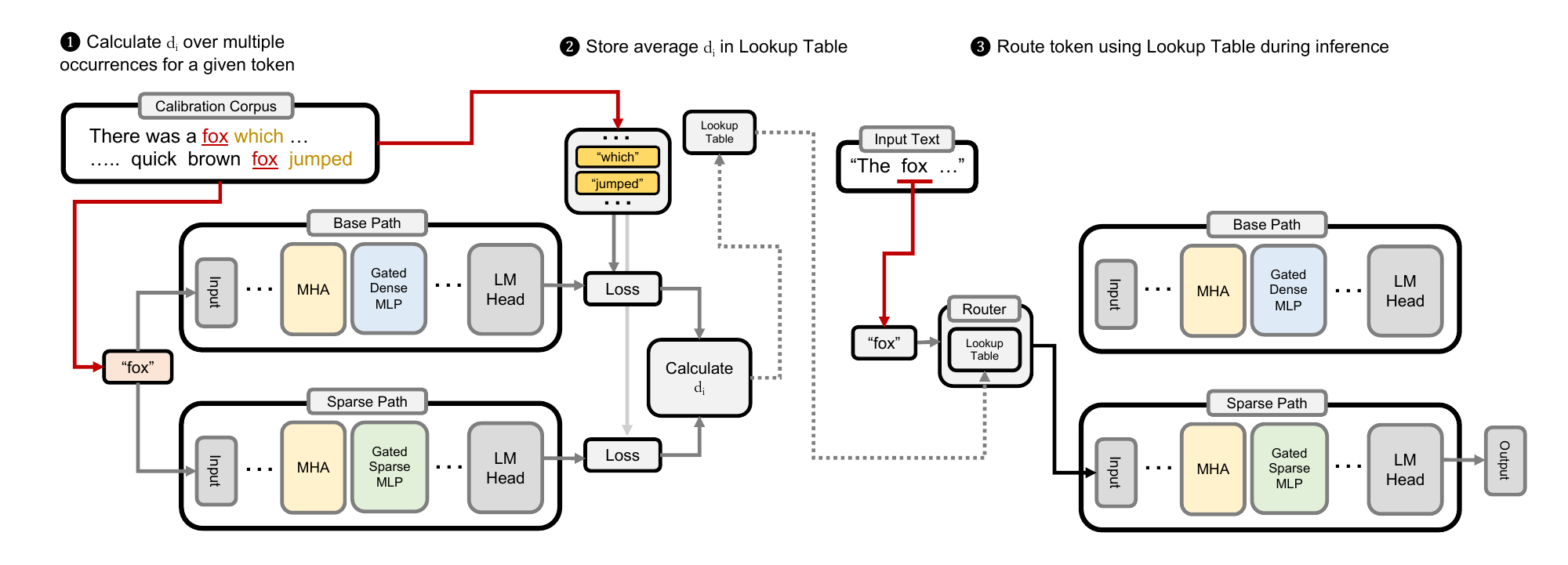}
    \caption{Overview of the proposed token routing method. For each token, $d_i$ is computed on a calibration corpus, and its average value is stored in a lookup table. During inference, the router uses this lookup table to pass the token to either the base path or the sparse path, depending on $\tau$. The base path and the sparse path use the same model weights, and only the MLP layer activation is adjusted.
    }
    \label{fig:token_routing}
\end{figure}
The activation thresholding methods discussed so far produce a single modified sparse model, which is then applied uniformly whenever the sparse configuration is used. While this static design is simple, it also makes the same sparse computation path for every token. Token routing asks whether this decision can be made more selectively. Instead of applying the sparse path uniformly, we allow each token to choose between the base model path, which preserves original model performance, and a sparse path, which applies activation thresholding for faster inference.

Considering practical model deployment scenarios, we only use the current token identity for the routing decision. This keeps the routing overhead minimal, since deployment requires only a lookup into a precomputed table. Under this constraint, the main goal is to assign each token a score that reflects how safe it is to use the sparse path on average.

To construct this score, we use a calibration corpus of tokenized sequences. For each token occurrence $i$ in that corpus, we run both the base path and the sparse path on the same token and record the next-token cross-entropy loss for that position. It is important to note that these losses are not intrinsic properties of the token type. Rather, they are measured from actual occurrences in the calibration corpus and depend on the surrounding sequence context. Routing calibration, therefore, builds a statistical estimate of how risky the sparse path would be for a given token.

Let $x_i$ denote the identity of the current token occurrence, and let $\ell_{\mathrm{base}}$ and $\ell_{\mathrm{sparse}}$ denote the per-token cross-entropy losses of the base path and sparse path, respectively. For the main routing target, we use the excess damage of the sparse path relative to the base path:
\begin{equation}
    d_i = \max(0, \ell_{\mathrm{sparse}} - \ell_{\mathrm{base}}) 
\end{equation}
This value is zero when the sparse path is no worse than the base path at occurrence $i$, and positive when the sparse path has higher loss. Therefore, this value approximates how harmful routing that token through the sparse path would be.

Since the routing mechanism only observes the current token identity $x_i$, it cannot predict the exact loss difference for each occurrence. Instead, it estimates a statistical average of this excess damage based on the calibration dataset:
\begin{equation}
    s(x) \approx \mathbb{E}[d_i \mid x_i = x]
\end{equation}
Lower scores indicate tokens that are, on average, safer to route through the sparse path.

We store these estimates in a lookup table over the vocabulary. Let $\mathcal{I}_x = \{i : x_i = x\}$ be the set of positions at which token $x$ appears in the routing corpus. The token score is estimated as
\begin{equation}
    \mathrm{route}(x_i; \tau) = \begin{cases}
        \text{sparse path} & \text{if } \hat{s}(x_i) < \tau, \\ 
        \text{base path} & \text{otherwise}
    \end{cases}
\end{equation}
For our routing method, a token sent to the sparse path uses the activation-thresholded MLP computation across transformer blocks, while tokens sent to the base path use the base MLP blocks. The threshold $\tau$ is used to control the routed-token fraction, and in turn, the quality-throughput trade-off.

As in SATS, calibration and operating-point selection are treated separately. We first estimate the token score table on a routing dataset, then calibrate the routing threshold on a held-out split. For a target routed fraction $r_{\mathrm{tar}}$, each candidate threshold $\tau$ induces a realized routed fraction $R(\tau)$ on the calibration split, and we select
\begin{equation}
    \tau^\star = \arg\min_{\tau} |R(\tau) - r_{\mathrm{tar}}|
\end{equation}
This produces different routing policies at different operating points. In our experiments, we report the performance of  25\%, 50\%, and 75\% target routing settings to validate the effectiveness of our approach. These routing settings are calibration operating points obtained by sweeping routing thresholds on a held-out calibration dataset and selecting the threshold whose realized routed fraction is closest to the corresponding target (e.g., 25\%, 50\%, or 75\% target sparsity).

\section{Experiments}
\label{sec:Experiments}
\textbf{Base Models:} We use the llama 3.1 8B~\citep{llama3} and Qwen 3 8B ~\citep{qwen3} models to demonstrate the effectiveness of our approaches.

\textbf{Sparsity Levels:} We evaluate the performance of CATS at 30\% and 50\% targeted sparsity levels, since at higher sparsity levels (e.g. 70\% or 90\%), too much information is removed, which makes them impractical for deployment. For proper comparison, we find the equivalent operating point that yields a similar realized sparsity level for SATS and compare the performance with CATS. 

For token routing experiments, we pick SATS matched to CATS 50\% for the two base models and observe their performance. Practically, at roughly 30\% sparsity level, both the accuracy loss and speedup gain are far too small to benefit from the quality-throughput trade-off offered by token routing, so we only demonstrate the effectiveness of the method at SATS matched to CATS 50\% sparsity.

For the final SATS configuration used in our experiments, we apply SATS to all MLP layers except the MLP in the final layer, where we retain the corresponding CATS threshold. Empirically, we found that thresholding this MLP layer had an outsized impact on text generation performance, and applying SATS there could disproportionately degrade language-modeling quality even when earlier layers might benefit from SATS. We therefore keep the CATS threshold in the final MLP layer as a pragmatic guardrail, while using SATS on the remaining layers to reach an equivalent CATS sparsity and demonstrate the effectiveness of our approach.

\textbf{Metrics}: We report the average accuracy over seven downstream tasks and perplexity on the Wiki-Text2 \citep{wikitext2} test dataset and a held-out test set from the RefinedWeb \citep{refinedweb} dataset. For throughput, we follow CATS and report the FP32 end-to-end throughput. We further conducted experiments to observe the variation in performance and latency between FP32 (full precision) and BF16 (half-precision).

\textbf{Datasets:} For downstream task evaluation, we use PIQA~\citep{piqa}, OpenBookQA~\citep{obqa}, ARC-Easy~\citep{arc}, Winogrande ~\citep{winogrande}, Hellaswag~\citep{hellaswag}, SciQ~\citep{sciq} and BoolQ ~\citep{boolq} from LM Eval Harness~\citep{eval-harness}. For fine-tuning base models, we used a small subset of the RefinedWeb dataset~\citep{refinedweb}.
Language modeling quality is measured with perplexity on a held-out split of RefinedWeb  and Wiki-Text2~\citep{wikitext2} test set. 

\textbf{Computational Environment}: For evaluating the quality and end-to-end throughput of the models, we utilized NVIDIA A100 40 GB. For fine-tuning base models using LoRA~\citep{lora}, we used NVIDIA RTX Pro 6000 96 GB. During inference, we used the transformers v4.57.6 Huggingface library, PyTorch v2.8, and CUDA v12.8. We used Triton for our GPU kernels, while torch.compile was used for evaluating end-to-end inference speedup.

\subsection{Comparison of SATS and CATS}

\begin{table}[h]
\centering
\footnotesize
\setlength{\tabcolsep}{4pt}
\renewcommand{\arraystretch}{1.2}
\begin{tabular}{l|ccccccc|cc}
\toprule 
Model & \begin{tabular}[c]{@{}c@{}}PIQA \\ acc $\uparrow$ \end{tabular} & \begin{tabular}[c]{@{}c@{}}OBQA \\ acc $\uparrow$\end{tabular} & \begin{tabular}[c]{@{}c@{}}Arc-E \\ acc $\uparrow$\end{tabular} & \begin{tabular}[c]{@{}c@{}}WG \\ acc $\uparrow$\end{tabular} & \begin{tabular}[c]{@{}c@{}}HS \\ acc $\uparrow$\end{tabular} & \begin{tabular}[c]{@{}c@{}}BoolQ \\ acc $\uparrow$\end{tabular} & \begin{tabular}[c]{@{}c@{}}SciQ \\ acc $\uparrow$\end{tabular} & \begin{tabular}[c]{@{}c@{}}Avg \\ acc $\uparrow$\end{tabular} & \begin{tabular}[c]{@{}c@{}}Realized. \\ Sparsity $\uparrow$\%\end{tabular} \\ \midrule
\textbf{llama 3.1 8B} & 79.27 & 45 & 82.32 & 74.11 & 79.36 & 83.09 & 96.6 & 77.11 & - \\
CATS 30\% & 79.11 & 43.8 & 81.23 & 73.56 & 79.28 & 82.72 & 96.7 & 76.63 & 26.32 \\
SATS (match CATS 30\%) & 78.89 & 45 & 81.48 & 73.09 & 79.34 & 82.66 & 96.5 & 76.71 & 26.52 \\
CATS 50\% & 78.56 & 44 & 79.59 & 70.96 & 78.13 & 80.64 & 96.2 & 75.44 & 43.9 \\
SATS (match CATS 50\%) & 79.11 & 44.4 & 80.05 & 73.88 & 78.46 & 80.89 & 96.6 & 76.2 & 43.9 \\ \midrule
\textbf{Qwen 3 8B} & 76.93 & 42.4 & 83.42 & 68.83 & 74.93 & 86.67 & 96.7 & 75.7 & - \\
CATS 30\% & 76.06 & 41.2 & 82.15 & 67.17 & 75.46 & 86.64 & 96.5 & 75.03 & 27.53 \\
SATS (match CATS 30\%) & 76.44 & 42.2 & 82.58 & 66.54 & 75.45 & 86.67 & 96.5 & 75.2 & 27.81 \\
CATS 50\% & 76.93 & 41.6 & 79.12 & 64.88 & 74.87 & 84.86 & 95.7 & 73.99 & 45.38 \\
SATS (match CATS 50\%) & 76.99 & 41 & 80.3 & 66.14 & 74.63 & 85.32 & 95.8 & 74.31 & 45.46 \\ \bottomrule
\end{tabular}
\caption{Downstream task performance of base models, CATS-based models with 30\% and 50\% targeted sparsity, and SATS-based models that match their respective realized sparsities. The SATS-based models (with matched realized sparsity comparable CATS) perform better in terms of average accuracy, which highlights the effectiveness of the calibration method.}
\label{tab:sats_v_cats_downstream}
\end{table}

\textbf{Training-free Performance: } The comparative performance of CATS~\citep{cats} and the proposed SATS is provided in Table \ref{tab:sats_v_cats_downstream} and Table \ref{tab:sats_v_cats_generation}, which demonstrate the downstream task performance and generation performance, respectively. 

The tables show that at the matched 30\% operating point, SATS is broadly competitive with CATS on both models. On llama 3.1 8B, SATS has a slight improvement in average downstream accuracy relative to CATS, but with a small perplexity cost. On Qwen 3 8B, SATS improves average accuracy and Wiki-Text2 perplexity, while RefinedWeb perplexity remains slightly worse. 

The clearer gains appear at the matched 50\% operating point. On llama 3.1 8B, SATS improves average downstream accuracy from 75.44 to 76.2 at an identical sparsity level, while providing better perplexity scores on both RefinedWeb and Wiki-Text2. On Qwen 3 8B, SATS improves average downstream accuracy from 73.99 to 74.31 while improving perplexity metrics. This demonstrates that at similar realized sparsity levels, the calibration method of SATS outperforms CATS in terms of performance. While we reported our accuracy and perplexity results in BF16, we also conducted experiments on the difference in performance between FP32 and BF16 on llama 3.1 8B on Table \ref{tab:perf_full_vs_half}, which demonstrates that the difference between the performance of the model in these data formats are minimal.

\begin{table}[h]
\centering
\footnotesize
\setlength{\tabcolsep}{4pt}
\renewcommand{\arraystretch}{1.2}
\begin{tabular}{lc|ccccc}
\toprule
\multicolumn{1}{c}{Model} & Dataset & Base Model & CATS 30\% & \begin{tabular}[c]{@{}c@{}}SATS \\ (match \\ CATS 30\%)\end{tabular} & CATS 50\% & \begin{tabular}[c]{@{}c@{}}SATS\\ (match \\ CATS 50\%)\end{tabular} \\ \midrule
\multirow{2}{*}{llama 3.1 8B} & WikiText-2 & 6.77 & 7.01 & 7.03 & 7.97 & 7.94 \\
 & RefinedWeb & 10.59 & 11.00 & 11.03 & 12.34 & 12.31 \\ \midrule
\multirow{2}{*}{Qwen 3 8B} & WikiText-2 & 10.52 & 10.24 & 10.18 & 10.63 & 10.4 \\
 & RefinedWeb & 14.99 & 14.57 & 14.64 & 15.25 & 15.02 \\ \bottomrule
\end{tabular}
\caption{Perplexity scores of the base models (lower is better), CATS at 30\% and 50\% target sparsity, and SATS (with matched realized sparsity comparable to CATS). At a higher sparsity level, SATS-based calibration outperforms CATS-based methods.}
\label{tab:sats_v_cats_generation}
\end{table}

\textbf{Fine-tuned model performance:} We compare the performance of the fine-tuned CATS and SATS on Table \ref{tab:sats_v_cats_downstream_finetuned} and Table \ref{tab:sats_v_cats_generation_fine_tuned}. While Table \ref{tab:sats_v_cats_downstream_finetuned} indicates that the difference between CATS and SATS is reduced in downstream task average accuracy, Table \ref{tab:sats_v_cats_generation_fine_tuned} highlights that for the generation task performance, SATS still performs better. 

While llama 3.1 8B models show little difference in downstream task performance compared to training-free methods, the effect of fine-tuning becomes more evident in language modeling tasks, where these models have lower levels of perplexity in general. It is also interesting to note that the fine-tuned Qwen 3 models outperformed even the corresponding base model.

\begin{table}[h]
\centering
\footnotesize
\setlength{\tabcolsep}{4pt}
\renewcommand{\arraystretch}{1.2}
\begin{tabular}{l|ccccccc|cc}
\toprule
Model & \begin{tabular}[c]{@{}c@{}}PIQA \\ acc $\uparrow$\end{tabular} & \begin{tabular}[c]{@{}c@{}}OBQA \\ acc $\uparrow$\end{tabular} & \begin{tabular}[c]{@{}c@{}}Arc-E \\ acc $\uparrow$\end{tabular} & \begin{tabular}[c]{@{}c@{}}WG \\ acc $\uparrow$\end{tabular} & \begin{tabular}[c]{@{}c@{}}HS \\ acc $\uparrow$\end{tabular} & \begin{tabular}[c]{@{}c@{}}BoolQ \\ acc $\uparrow$\end{tabular} & \begin{tabular}[c]{@{}c@{}}SciQ \\ acc $\uparrow$ \end{tabular} & \begin{tabular}[c]{@{}c@{}}Avg \\ acc $\uparrow$\end{tabular} & \begin{tabular}[c]{@{}c@{}}Realized \\ Sparsity\% $\uparrow$\end{tabular} \\ \midrule
\textbf{llama 3.1 8B} & 79.27 & 45 & 82.32 & 74.11 & 79.36 & 83.09 & 96.6 & 77.11 & - \\
CATS 50\% & 78.94 & 42.2 & 79.59 & 72.69 & 78.09 & 81.22 & 96.8 & 75.65 & 44.6 \\
SATS (match CATS 50\%) & 78.78 & 43.2 & 79.59 & 71.98 & 78.09 & 81.47 & 96.4 & 75.64 & 44.4 \\ \midrule
\textbf{Qwen 3 8B} & 76.93 & 42.4 & 83.42 & 68.83 & 74.93 & 86.67 & 96.7 & 75.7 & - \\
CATS 50\% & 78.18 & 41.4 & 82.53 & 70.24 & 75.29 & 85.29 & 96.6 & 75.65 & 45.2 \\
SATS (match CATS 50\%) & 78.18 & 42.8 & 81.99 & 71.35 & 75.54 & 85.44 & 96.5 & 75.97 & 45.7 \\ \bottomrule
\end{tabular}
\caption{Downstream task performance of base models, \textbf{fine-tuned} CATS-based models with 50\% targeted sparsity, and \textbf{fine-tuned} SATS-based models that match its realized sparsities. }
\label{tab:sats_v_cats_downstream_finetuned}
\end{table}

\begin{table}[h]
\centering
\footnotesize
\renewcommand{\arraystretch}{1.15}
\setlength{\tabcolsep}{5pt}
\begin{tabular}{lc|ccc}
\toprule
\multicolumn{1}{c}{Model} & Dataset & Base Model & CATS 50\% & \begin{tabular}[c]{@{}c@{}}SATS\\ (match CATS 50\%)\end{tabular} \\ \midrule
\multirow{2}{*}{llama 3.1 8B} & WikiText-2 & 6.77 & 7.6 & 7.57 \\
 & RefinedWeb & 10.59 & 11.4 & 11.37 \\ \midrule
\multirow{2}{*}{Qwen 3 8B} & WikiText-2 & 10.52 & 8.47 & 8.41 \\
 & RefinedWeb & 14.99 & 12.48 & 12.38 \\ \bottomrule
\end{tabular}
\caption{Perplexity scores of the base models (lower is better), \textbf{fine-tuned} CATS at 50\% target sparsity, and \textbf{fine-tuned} SATS (with matched realized sparsity comparable to CATS).}
\label{tab:sats_v_cats_generation_fine_tuned}
\end{table}

\begin{table*}[]
\footnotesize
\renewcommand{\arraystretch}{1.2}
\setlength{\tabcolsep}{5pt}
\centering
\begin{tabular}{c|c|c|c|c|c}
\hline
Model & Quantization & Avg Acc. & \begin{tabular}[c]{@{}c@{}}Avg. \\ Sparsity\%\end{tabular} & \begin{tabular}[c]{@{}c@{}}WT2 \\ PPL\end{tabular} & \begin{tabular}[c]{@{}c@{}}RW\\ PPL\end{tabular} \\ \hline
\multirow{2}{*}{Baseline} & bf16 & 77.11 & - & 6.77 & 10.59 \\
 & fp32 & 77.21 & - & 6.76 & 10.58 \\ \hline
\multirow{2}{*}{CATS 50\%} & bf16 & 75.44 & 43.9 & 7.97 & 12.34 \\
 & fp32 & 75.69 & 43.8 & 7.95 & 12.33 \\ \hline
\multirow{2}{*}{\begin{tabular}[c]{@{}c@{}}SATS (Match \\ CATS 50\%)\end{tabular}} & bf16 & 76.2 & 43.9 & 7.94 & 12.31 \\
 & fp32 & 76 & 43.7 & 7.93 & 12.28 \\ \hline
\end{tabular}
 \caption{Comparison of performance with full precision and half-precision on llama 3.1 8B}
\label{tab:perf_full_vs_half}
\end{table*}

\subsection{Evaluation of Token Routing}
We use the SATS matched to CATS-50\% operating point and evaluate the effectiveness of the token routing method on both training-free and fine-tuned settings. 

\textbf{Training-free Methods:} The results in Table \ref{tab:token_routing} indicate that token routing improves the quality relative to the static SATS method while preserving most of the throughput gain over dense baseline inference. On llama 3.1 8B, the 50\% target routing provides the best balance between quality and speed in terms of accuracy and perplexity. 

On Qwen 3 8B, the 50\% target routing improves average accuracy from 74.31 to 75.11, while improving the perplexity metrics. Though static SATS provides the best throughput gain, routed SATS remain fast while boosting average accuracy. While we have measured our latency in FP32 throughout the experiments, we also observed the BF16 performance of the proposed token routing method on Table \ref{tab:token_routing_half_precision} for llama 3.1 8B, which follows similar pattern as the FP32 latency performance of the model.

\textbf{Fine-tuning Methods:} The routing results for the fine-tuned models are shown in Table \ref{tab:token_routing_finetuned}. The speedup gain in the fine-tuned routed models is less pronounced compared to the training-free setting, since the sparse path needs to perform both the thresholded base MLP computation and the LoRA update. In our initial experiments, the naive implementation introduced substantial extra overhead from separate low-rank operations. To mitigate this cost, we use LoRA-aware sparse kernels that combine selected LoRA computations with the sparse MLP path and reduce kernel-launch and small-matrix overhead through custom Triton kernels. Though this additional computation narrows the relative gain, the routed fine-tuned models still retain a clear throughput advantage over dense inference. 

After fine-tuning the models, the sparse model performance improves notably for Qwen 3. It is interesting to note that with the improved performance of the fine-tuned Qwen 3 model, routing the tokens to the SATS path more (e.g., 50\% or 75\% route target) improves the accuracy, even surpassing the performance of both static dense and sparse models in terms of average accuracy.
\begin{table}[h]
\centering
\footnotesize
\renewcommand{\arraystretch}{1.2}
\setlength{\tabcolsep}{5pt}
\begin{tabular}{c|c|c|cccc}
\toprule
Model & Method & \begin{tabular}[c]{@{}c@{}}Target \\ Route\%\end{tabular} & Avg. Acc & \begin{tabular}[c]{@{}c@{}}RW\\ PPL\end{tabular} & \begin{tabular}[c]{@{}c@{}}WT2\\ PPL\end{tabular} & tokens/Sec \\ \midrule
\multirow{5}{*}{llama 3.1 8B} & Baseline & 0 & 77.11 & 10.59 & 6.77 & 32.12 \\ \cline{2-7} 
 & \multirow{3}{*}{Routed SATS} & 25 & 77.03 & 11.04 & 7.01 & 31.88 \\
 &  & 50 & 76.45 & 11.43 & 7.24 & 35.41 \\
 &  & 75 & 76.02 & 11.86 & 7.51 & 35.68 \\\cline{2-7} 
 & Static SATS & 100 & 76.2 & 12.31 & 7.94 & 37.08 \\ \midrule
\multirow{5}{*}{Qwen 3 8B} & Baseline & 0 & 75.7 & 14.99 & 10.52 & 31.07 \\ \cline{2-7} 
 & \multirow{3}{*}{Routed SATS} & 25 & 75.19 & 14.96 & 10.45 & 34.4 \\
 &  & 50 & 75.11 & 14.91 & 10.39 & 35.03 \\
 &  & 75 & 74.81 & 14.89 & 10.36 & 35.24 \\ \cline{2-7} 
 & Static SATS & 100 & 74.31 & 15.02 & 10.4 & 35.59 \\ \bottomrule
\end{tabular}
\caption{Downstream task accuracy and language modeling performance of base model, token routed SATS, and static SATS (both use the SATS matched to CATS 50\% target sparsity). The target route percentage is measured from a calibration dataset. The results indicate that the token routing offers higher accuracy while offering faster throughput over the baseline dense model.}
\label{tab:token_routing}

\end{table}
\begin{table}[h]
\centering
\footnotesize
\renewcommand{\arraystretch}{1.2}
\setlength{\tabcolsep}{5pt}
\begin{tabular}{c|c|c|cccc}
\toprule
Model & Method & \begin{tabular}[c]{@{}c@{}}Target \\ Route\%\end{tabular} & Avg. Acc & \begin{tabular}[c]{@{}c@{}}RW\\ PPL\end{tabular} & \begin{tabular}[c]{@{}c@{}}WT2\\ PPL\end{tabular} & tokens/Sec \\ \midrule
\multirow{5}{*}{llama 3.1 8B} & Baseline & 0 & 77.11 & 10.59 & 6.77 & 32.12 \\ \cline{2-7} 
 & \multirow{3}{*}{Routed SATS} & 25 & 76.89 & 10.78 & 6.93 & 34.54 \\
 &  & 50 & 76.5 & 10.95 & 7.11 & 35.19 \\
 &  & 75 & 76.06 & 11.19 & 7.25 & 34.95 \\ \cline{2-7} 
 & Static SATS & 100 & 75.79 & 11.37 & 7.57 & 35.7 \\ \midrule
\multirow{5}{*}{Qwen 3 8B} & Baseline & 0 & 75.7 & 14.57 & 10.24 & 31.07 \\ \cline{2-7} 
 & \multirow{3}{*}{Routed SATS} & 25 & 75.64 & 13.87 & 9.52 & 32.76 \\
 &  & 50 & 76.01 & 13.06 & 8.94 & 32.63 \\
 &  & 75 & 76.31 & 12.61 & 8.72 & 33.8 \\ \cline{2-7} 
 & Static SATS & 100 & 75.97 & 12.38 & 8.41 & 34 \\ \bottomrule
\end{tabular}
 \caption{Downstream task accuracy and language modeling performance of base model, token routed SATS \textbf{fine-tuned} and static SATS \textbf{fine-tuned} (both use the SATS matched to CATS 50\% target sparsity). The target route percentage is measured from a calibration dataset. The results indicate that the token routing offers higher accuracy while offering faster throughput over the baseline dense model.}
\label{tab:token_routing_finetuned}
\end{table}

\begin{table*}[]
\centering
\footnotesize
\renewcommand{\arraystretch}{1.2}
\setlength{\tabcolsep}{5pt}
\begin{tabular}{c|c|c|c}
\hline
Model & Method & \begin{tabular}[c]{@{}c@{}}Target\\  Route\%\end{tabular} & Tokens/Sec \\ \hline
\multirow{5}{*}{llama 3.1 8B} & Baseline & 0 & 52.2 \\ \cline{2-4} 
 & \multirow{3}{*}{Routed SATS} & 25 & 50.2 \\
 &  & 50 & 54.1 \\
 &  & 75 & 54.3 \\ \cline{2-4} 
 & Static SATS & 100 & 57 \\ \hline
\end{tabular}
 \caption{BF16 (half-precision) latency of base llama 3.1 8B model, token routed SATS and static SATS (both use the SATS matched to CATS 50\% target sparsity)}
\label{tab:token_routing_half_precision}
\end{table*}

\section{Discussion and Conclusion}
We studied how to utilize sparsity in LLMs through sensitivity-aware activation thresholding and token-level routing. Our results indicate that the proposed SATS improves over percentile-based calibration at similar sparsity levels. We also demonstrated that token routing improves downstream accuracy and language-modeling quality while remaining faster than dense inference.

These results indicate that both activation-level and token-level sparsity can be utilized for efficient LLM inference. Better threshold calibration improves the static sparse model quality, while token routing allows selectively applying the sparse model for a better quality-throughput trade-off. Overall, the proposed SATS and token routing improve practical LLM inference efficiency.

\bibliographystyle{plainnat}
\bibliography{refs}

\end{document}